\documentclass[
]{ceurart}

\usepackage{listings}
\begin{document}

\copyrightyear{2023}
\copyrightclause{Copyright for this paper by its authors.
  Use permitted under Creative Commons License Attribution 4.0
  International (CC BY 4.0).}

\conference{CLEF 2023: Conference and Labs of the Evaluation Forum, 
    September 18--21, 2023, Thessaloniki, Greece}


\title{Overview of PlantCLEF 2023: Image-based Plant Identification at Global Scale}

\author[1]{Herv\'e Go\"eau}[%
orcid=0000-0003-3296-3795,
email=herve.goeau@cirad.fr
]

\author[1]{Pierre Bonnet}[%
orcid=0000-0002-2828-4389,
email=pierre.bonnet@cirad.fr
]

\author[2]{Alexis Joly}[%
orcid=0000-0002-2161-9940,
email=alexis.joly@inria.fr
]
\address[1]{CIRAD, UMR AMAP, Montpellier, Occitanie, France}
\address[2]{Inria, LIRMM, Univ Montpellier, CNRS, Montpellier, France}

\begin{abstract}
The world is estimated to be home to over 300,000 species of vascular plants. In the face of the ongoing biodiversity crisis, expanding our understanding of these species is crucial for the advancement of human civilization, encompassing areas such as agriculture, construction, and pharmacopoeia. However, the labor-intensive process of plant identification undertaken by human experts poses a significant obstacle to the accumulation of new data and knowledge. Fortunately, recent advancements in automatic identification, particularly through the application of deep learning techniques, have shown promising progress. Despite challenges posed by data-related issues such as a vast number of classes, imbalanced class distribution, erroneous identifications, duplications, variable visual quality, and diverse visual contents (such as photos or herbarium sheets), deep learning approaches have reached a level of maturity which gives us hope that in the near future we will have an identification system capable of accurately identifying all plant species worldwide. The PlantCLEF2023 challenge aims to contribute to this pursuit by addressing a multi-image (and metadata) classification problem involving an extensive set of classes (80,000 plant species). This paper provides an overview of the challenge's resources and evaluations, summarizes the methods and systems employed by participating research groups, and presents an analysis of key findings.
\end{abstract}

\maketitle

\section{Introduction}

The world is home to an estimated 300,000 species of vascular plants, and the discovery and description of new plant species continue to occur each year \cite{christenhusz2016number}. The remarkable diversity of plants has played a pivotal role in the advancement of human civilization, providing resources such as food, medicine, building materials, recreational opportunities, and genetic reservoirs \cite{naeem2009biodiversity}. Moreover, plant diversity plays a crucial role in maintaining the functioning and stability of ecosystems \cite{naeem2009biodiversity}. However, our understanding of plant species remains limited. For the majority of species, we lack knowledge about their specific roles within ecosystems and their potential utility to humans. Additionally, information regarding the geographic distribution and population abundance of most species remains scarce \cite{cronk2016plant}.

Over the past two decades, the biodiversity informatics community has made significant efforts to develop global initiatives, digital platforms, and tools to facilitate the organization, sharing, visualization, and analysis of biodiversity data \cite{parr2014encyclopedia, wheeler2004if}. Nonetheless, the process of systematic plant identification poses a significant obstacle to the aggregation of new data and knowledge at the species level. Botanists, taxonomists, and other plant experts spend substantial time and energy on species identification, which could be better utilized in analyzing the collected data. 

As previously discussed by \cite{Gaston29042004}, the routine identification of previously described species shares similarities with other human activities that have successfully undergone automation. In recent years, automated identification has made significant advancements, particularly due to the development of deep learning techniques, thanks to the rise of Convolutional Neural Networks (CNNs) \cite{lecun2015deep}. The long-term evaluation of automated plant identification, conducted as part of the LifeCLEF initiative \cite{joly2019biodiversity}, demonstrates the impact of CNNs on performance within a few years. In 2011, the best evaluated system achieved a mere 57\% accuracy on a straightforward classification task involving only 71 species captured under highly uniform conditions (scans or photos of leaves on a white background). In contrast, by 2017, the best CNN achieved an 88.5\% accuracy on a far more complex task encompassing 10,000 plant species, characterized by imbalanced, heterogeneous, and noisy visual data \cite{plantclef2017}. Moreover, in 2018, the best system outperformed five out of nine specialists in re-identifying a subset of test images \cite{expertclef2018}.

Existing plant identification applications, due to their growing popularity, present opportunities for high-throughput biodiversity monitoring and the accumulation of specific knowledge \cite{nugent2018inaturalist,waldchen2018automated,affouard2017pl}. However, they often face the challenge of being restricted to specific regional floras or limited to the most common species. With an increasing number of species exhibiting a transcontinental range, such as naturalized alien species \cite{van2019global} or cultivated plants, relying on regional floras for identification becomes less reliable. Conversely, focusing solely on the most prevalent species disregards the broader implications for biodiversity.

To address these challenges, during two years of competition, the PlantCLEF 2022 and 2023 challenges introduced a multi-image (and metadata) classification problem involving an extensive number of classes, specifically 80,000 plant species. Convolutional Neural Networks (CNNs) and the recent Vision Transformers (ViTs) techniques emerge as the most promising solutions for tackling such large-scale image classification tasks. However, previous studies had not reported image classification results of this magnitude, regardless of whether the entities were biological or not. This paper presents the challenge's resources and evaluations, summarizes the approaches and systems employed by participating research groups.

\section{Dataset}

To thoroughly evaluate the aforementioned scenario on a large scale and in realistic conditions, two distinct training datasets were developed and shared: the "trusted" dataset and the "web" dataset. These datasets encompassed a total of 4 million images across 80,000 plant species, sourced from various origins.\\
\\
\textbf{"Trusted" training set:} this training dataset is based on a carefully curated selection of more than 2.9M images covering 80k plant species aggregated, shared and collected mainly by GBIF (Global Biodiversity Information Facility). This type of data is aggregated from academic sources such as museums, universities, and national institutions, as well as collaborative platforms like inaturalist, and Pl@ntNet, implying a fairly high certainty of determination quality. We initially formed an extensive dataset using the GBIF portal, which includes nearly 16 million occurrences of vascular plants (Tracheophyta) comprising ferns, conifers, and flowering modern plants \cite{https://doi.org/10.15468/dl.ej7kn5}. This initial selection, however, exhibited significant imbalance, with some species having tens of thousands of images while others had only one. To ensure class equilibrium and prevent dataset inflation, we limited the number of images per species to approximately 100. The selected images focus on views that are optimal for plant identification, such as close-ups of flowers, fruits, leaves, and trunks.\\
\\
\textbf{"Web" training set:} in contrast, the "web" training dataset was compiled from a collection of web images obtained from search engines like Google and Bing. This initial collection contained millions of images, but it suffered from significant errors in species identification, a high presence of duplicate images, and a large number of images that were less suitable for visual plant identification, such as herbarium images, landscapes, microscopic views, and unrelated subjects. To address these issues, a semi-automatic revision process was conducted to minimize the number of irrelevant images and maximize the inclusion of close-ups of relevant plant features. The "web" dataset ultimately consisted of approximately 1.1 million images, covering around 57,000 plant species.\\
\\
\textbf{Test set:} For the evaluation of the models, a separate test set was constructed using multi-image plant observations collected on the Pl@ntNet platform throughout 2021, ensuring that they were not present in the training datasets. Only observations with a high confidence score, determined through the collaborative review process on Pl@ntNet, were selected for the challenge, ensuring a high level of determination quality. The review process involved individuals with varying levels of expertise, ranging from beginners to world-leading experts, with different weights given to their judgments. The test set consisted of approximately 27,000 plant observations, comprising around 55,000 images related to approximately 7,300 plant species.\\
\\
Table\ref{tab:datastats} presents various statistics about the three datasets. One notable observation is the significant difference in the number of species between the training sets and the test set. This difference primarily stems from the challenge of collecting a large amount of expert-verified data from botanists on such a scale. However, this difference aligns with the realistic scenario faced by automatic identification systems like Pl@ntNet and Inaturalist. These systems need to be capable of recognizing a wide range of species without prior knowledge of which species will be frequently requested or completely overlooked. This characteristic reflects the goal of these systems to identify as many species as possible and adapt to unpredictable user requests.

\begin{table}[!h]
    \caption{Statistics of the LifeCLEF 2023 Plant Identification Task: "n/s" means not specified }
    \centering
    \begin{tabular}{|r|r|r|r|r|r|r|}
    \hline
    Dataset         & Images    & Observations  & Classes (species) & Genera    & Families  & Orders \\
    \hline
    Train "trusted"	& 2,886,761 & n/s           & 80,000	        & 9,603     & 483       & 84 \\
    Train "web"     & 1,071,627 & n/s           & 57,314            & 8,649     & 479       & 84 \\
    Test            & 55,307    & 26,869        & 7,339             & 2,527     &           &  \\
    \hline
\end{tabular}
\label{tab:datastats}
\end{table}

\section{Task Description}

The challenge was hosted during two years as two rounds in the AICrowd plateform\footnote{\url{https://www.aicrowd.com/challenges/lifeclef-2022-23-plant}}. The task was evaluated as a plant species retrieval task based on multi-image plant observations from the test set. The goal was to retrieve the correct plant species among the top results of a ranked list of species returned by the evaluated system. During the first year of competition in 2022, the participants had access to the training set in mid-February 2022, the test set was published 6 weeks later in early April, and the round of submissions was then open during 5 weeks. During the second round in 2023, the training and test data remained exactly the same (the ground truth on the test set being kept secret). The submission system remained open from mid-March to mid-May. 

The metric used for the evaluation of the task is the Macro Average (by species) Mean Reciprocal Rank (MA-MRR). The Mean Reciprocal Rank (MRR) is a statistic measure for evaluating any process that produces a list of possible responses to a sample of queries ordered by probability of correctness. The reciprocal rank of a query response is the multiplicative inverse of the rank of the first correct answer. The MRR is the average of the reciprocal ranks for the whole test set:
\begin{equation}
    MRR = \frac{1}{O} \sum_{i=1}^O \frac{1}{\text{rank}_i}
\end{equation} 
where $O$ is the total number of plant observations (query occurrences) in the test set and $\text{rank}_i$ is the rank of the correct species of the plant observation $i$.\\

However, the Macro-Average version of the MRR (average MRR per species in the test set) was used because of the long tail of the data distribution to rebalance the results between under- and over-represented species in the test set:
\begin{equation}
    MA-MRR = \frac{1}{S} \sum_{j=1}^S \frac{1}{O_j} \sum_{i=1}^{O_j} \frac{1}{\text{rank}_i}
\end{equation} 
where $S$ is the total number of species in the test set, $O_j$ is the number of plant observations related to a species $j$.

\section{Participants and methods}

During the two years of the challenge, a total of 195 people expressed an interest in signing up for the challenge. Among this large raw audience, 8 research groups finally succeeded in submitting run files (8 the first year and 3 the second year). Details of the used methods and evaluated systems during the 2023 round are synthesized below and further developed in the working notes of the participants (Mingle Xu \cite{MingleXu2023}, Neuon AI\cite{NeuonAI2023}. Table \ref{tab:rawresults} reports the results while describing in various columns the main characteristics that distinguish each method from the others: type of architecture, training set used, pre-training method, taxonomic levels used.  Complementary, the following paragraphs give a few more details about the methods and the overall strategy employed by each participant (the paragraphs are sorted in descending order of the best score obtained by each team).\\
\\
\textbf{Mingle Xu, South Korea, 9 runs, \cite{MingleXu2023}}: the team's work is founded on the utilization of a Vision Transformer (ViT) that has been pre-trained using a Self Supervised Learning (SSL) technique, which is a recent and increasingly popular approach in the field of computer vision. This approach is quite disruptive as it deviates from the traditional Supervised Transfer Learning (STL) method. Typically, in a STL approach, a neural network is initially trained from scratch using labeled data for a classification task on a generic dataset such as ImageNet 1k or 22k, and the network is then subsequently fine-tuned on a specific dataset that possesses a distinct set of labels. In contrast, Self Supervised Learning (SSL) methods operate without the need for labeled data. The premise is that a network pre-trained with an SSL method can extract superior features that exhibit improved generalization abilities. These extracted features can subsequently be fine-tuned in a supervised manner, enabling their effective utilization for diverse downstream tasks, including image classification and object detection. In contemporary times, the primary focus of state-of-the-art methods lies not in devising new architectures, as ViT has emerged as the default choice. Instead, the emphasis is on determining the optimal SSL approach for pre-training a ViT model.
As an example, in the previous year, the team led by Mingle Xu achieved the best results using a pre-trained ViT-large model based on a Masked Auto-Encoder (MAE) approach \cite{he2022masked}. They obtained a remarkable MA-MRR score of 0.64079 (post-challenge). The concept of MAE draws inspiration from the successful masked language modeling technique commonly used in Natural Language Processing, notably popularized by BERT \cite{devlin2018bert}. The process of masking data was challenging to apply to CNN-based architectures, whereas it becomes relatively straightforward with vision transformers since they operate internally using visual patches or "tokens" along with positional embedding. MAE shares similarities with BEIT \cite{bao2021beit}, wherein the self-supervised task involves training a backbone vision transformer to predict missing tokens from partially masked images.

During this year's participation, Mingle Xu explored declinations of runs based on the vision-centric foundation model EVA \cite{fang2022eva}, that was the state-of-the-art position during the challenge in the first quarter of 2023. EVA is a pretraining strategy that combines CLIP\cite{clip} and MVP\cite{mvp}. CLIP maximizes the relationships between paired text and images, while MVP integrates MAE and CLIP to enhance pretraining. MVP freezes the CLIP image encoder and trains the vision part using a loss function that minimizes the distance between frozen CLIP features and vision model features. EVA scales up MVP by using larger models and more datasets, resulting in improved performance across various tasks. Overall, EVA leverages multimodal information and scalability for better semantic learning.

Mingle Xu conducted an investigation into the finetuning of pre-trained EVA models using various approaches. This included species ablations with limited image data (runs 1, 2, 4, 6), augmenting the "trusted" training set with additional images from the "web" training set (runs 8, 9, 10), starting from a self-supervised learning (SSL)-only pre-trained model (run 3), or employing intermediate supervised finetuning on ImageNet 22k (all other runs). The best run (MingleXuRun 8) reached an impressive MA-MRR of 0.67395.\\
\\
\textbf{Neuon AI, Malaysia, 10 runs, \cite{NeuonAI2023}}: this participant used various ensembles of models finetuned using most of the time all the training dataset available ("trusted" and "web") training datasets, mainly based on the Inception-ResNet-v2 architectures \cite{szegedy2016inception} (and on a Inception-v4 to a lesser extent). All the models are directly finetuned CNNs but as a multi-task classification related to five taxonomy levels (species, genus, family, order and "class" in the botanical sense), instead of the default species level. All the runs were finetuned . They then explored various ways to improve the performances: more data augmentation, a balanced batching method, a multi-organ and single-organ training scheme, and finally a feature embedding comparison instead of the traditional softmax function. 
The same data augmentation techniques were applied for all runs and included random cropping, horizontal flipping, color distortion, bi-cubic resizing, random hue and random contrast. The balanced batching method consisted to limit the selection of training images for a species in an epoch to a maximum of 16 samples in order to avoid any bias towards any particular species and preventing poor performance on underrepresented species (runs 2 and 4). A multi-organ training scheme was used for run 4: the approach involved training multiple models on smaller sub-datasets that exclusively consisted of images tagged with either the Flower, Bark, Fruit, Habit, or Leaf tag. 
The feature embedding comparison was used in runs 3, 6 and 8. It relies on calculating distances between test and training images, utilizing cosine similarity applied to the feature vectors of a single test image and all the training images. The feature embedding are directly the features extracted from the model before the fully-connected last layer. Distances scores are transformed into probabilities using Inverse Distance Weighting, allowing for class ranking, and the class with the highest probability is finally representing the most confident prediction.

\section{Results}

\begin{table}[h]
    \caption{Results of the LifeCLEF 2023 Plant Identification Task. \textbf{Archi. (architecture):} IRv2: Inception-ResNet-v2, Iv4: Inception-v4, ViT-L: Vision Transformer Large,  ViT-B: Vision Transformer Base. \textbf{Datasets:} IN1k: ImageNet1k, IN21k: ImageNet21k, PlantCLEF2023: T (Trusted), W (Web), TW (Trusted \& Web), TWO (Trusted \& Web by Organ), T@N (Trusted with classes having more than N images). \textbf{Pre-training methods:} STL: Supervised Transfer Learning., SSL: Self Supervised Learning, EVA: Explore the limits of Visual representation at scAle, MAE: Masked Auto-Encoder . \textbf{Taxo. (Taxonomy):} Sp. (species), All (species, genus, family, order, class). \textbf{Org. (Organs):} M (Multi-organs), S (Single-organ)).}
    \centering
    \begin{tabular}{|l|c|c|c|c|c|c|}
    \hline
    Team run name   & Archi         & Pre-training  & Train     & Taxo.    & Org.   & MA-MRR\\
    \hline
    MingleXuRun8    & ViT-L         & SSL EVA IN21k -> STL IN21k & TW        & Sp.      & M      & 0.67395 \\ 
    MingleXuRun9    & ViT-L         & SSL EVA IN21k -> STL IN21k & TW        & Sp.      & M      & 0.66330 \\ 
    MingleXuRun10	& ViT-L         & SSL EVA IN21k -> STL IN21k & TW        & Sp.      & M      & 0.65695 \\ 
    MingleXuRun5	& ViT-L         & SSL EVA IN21k -> STL IN21k & T        & Sp.      & M      & 0.65035 \\ 

    MingleXuRun3	& ViT-L         & SSL EVA IN21k & T        & Sp.      & M      & 0.64871 \\ 
    
    MingleXuRun6	& ViT-L         & SSL EVA IN21k -> STL IN21k & T@7        & Sp.      & M      & 0.64201 \\ 
    NeuonAIRun9		& Iv4, IRv2     & STL IN1k      & TW        & All      & M      & 0.61813 \\ 
    NeuonAIRun7		& Iv4, IRv2     & STL IN1k      & TW,T      & All      & M      & 0.61561 \\ 
    NeuonAIRun10	& IRv2          & STL IN1k      & TW,T      & All      & M      & 0.61406 \\ 
    MingleXuRun2	& ViT-L         & SSL EVA IN21k -> STL IN21k & T@36        & Sp.      & M      & 0.57514 \\ 
    NeuonAIRun5		& IRv2          & STL IN1k      & TW        & All      & M      & 0.55040 \\ 
    MingleXuRun4	& ViT-L         & SSL EVA IN21k -> STL IN21k & T@50        & Sp.      & S      & 0.54846 \\ 
    NeuonAIRun1		& IRv2          & STL IN1k      & TW        & All      & M      & 0.54242 \\ 
    NeuonAIRun2		& IRv2          & STL IN1k      & TW        & All      & M      & 0.46606 \\ 
    NeuonAIRun6		& IRv2          & STL IN1k      & TW        & All      & M      & 0.46476 \\ 
    NeuonAIRun8		& IRv2          & STL IN1k      & TW        & All      & M      & 0.45910 \\ 
    NeuonAIRun3		& IRv2          & STL IN1k      & TW        & All      & M      & 0.45242 \\ 
    NeuonAIRun4		& IRv2          & STL IN1k      & TWO       & All      & S      & 0.33926 \\ 
    MingleXuRun1	& ViT-L         & SSL EVA IN21k -> STL IN21k & T@100      & Sp.      & M      & 0.33239 \\ 
    \hline
    BestRun2022     & ViT-L         & SSL MAE IN1k  & T         & Sp.      & M      & 0.64079 \\

\hline
\end{tabular}
\label{tab:rawresults}
\end{table}

\begin{figure}[t]
\centering
\includegraphics[width=0.95\linewidth]{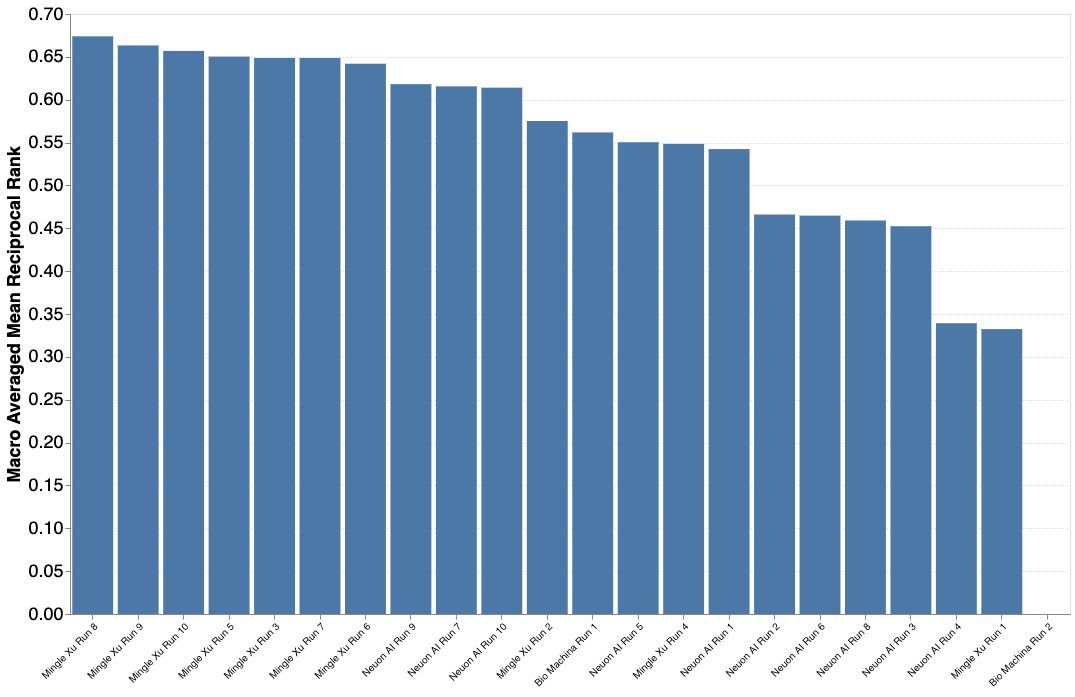}
\caption{Scores achieved by all systems evaluated within the plant identification task of LifeCLEF 2023}
\label{fig:PlantCLEF2023OfficialScore}
\end{figure}

We report in Figure \ref{fig:PlantCLEF2023OfficialScore} the performance achieved by the collected runs. Table \ref{tab:rawresults} provides the results achieved by each run as well as a brief synthesis of the methods used in each of them.\\
\\
\textbf{ViT SSL are better than CNN STL}: The most impressive outcomes were achieved by vision transformer-based approaches, particularly the vision-centric foundation model EVA \cite{fang2022eva}, that was the state-of-the-art position during the challenge in the first quarter of 2023. While CNN-based approaches also produced respectable results, with a maximum MA-MRR of 0.61813 (NeuonAIRun9), they still fell notably short of the highest score attained by an EVA approach. The best EVA approach, achieved a remarkable MA-MRR of 0.67395 (MingleXuRun8).\\
\\
\textbf{The noisy web training dataset helps}: incorporating the comprehensive PlantCLEF training dataset, which includes both the trusted and web datasets, yielded notable benefits despite the extended training duration and the inherent residual noise present in the web dataset. The inclusion of the web training dataset led to a significant improvement in performance, as evidenced by the MA-MRR reaching 0.67395 (MingleXuRun8), surpassing the maximum of 0.65035 (MingleXuRun5) achieved without its incorporation. However, it's possible that the web dataset has been well curated and that the noise level isn't as high as one might think.\\
\\
\textbf{Species ablation was not relevant}: the reduction of the training set by removing the classes with the fewest images (MingleXuRun1-4-2-6 vs 5) implies a significant drop in performance. This observation highlights a crucial point: the presence of a direct correlation between training and test data is not always guaranteed. It underlines the importance of including all classes, including those associated with uncommon species, to meet the challenge of monitoring plant biodiversity. By including a diverse range of classes, even those associated with less common species, we can better grasp the true extent and variability of plant life. This holistic approach ensures a more promising understanding of plant biodiversity for effective monitoring and conservation efforts in the future. It is a reminder that comprehensive and inclusive datasets are essential for accurate and reliable analysis in the field of plant biodiversity.\\
\\
\textbf{Combining models dedicated to specific organs deteriorates results}: Intuitively, one might say that it's interesting to specialize models on organ-based learning subsets, as botanists eventually learn to analyze organ structure and appearance independently and separately. However, a noteworthy observation is that one of the poorest results in the challenge emerged when combining models trained on specific organ sub-datasets (NeuonAIRun4 with a MA-MRR of 0.33926). A possible explanation for this outcome can be attributed to the loss of a significant number of species per organ. Statistics reported by the authors indicate that by focusing solely on fruits, for instance, the resulting dataset encompasses only approximately 27,000 species, significantly lower than the total 80,000 species available. This reduction in species coverage likely hampers the model's ability to generalize and accurately classify a broader range of plant organisms.

\section{Conclusion}


This paper presented the overview and the results of the LifeCLEF 2023 plant identification challenge following the 12 previous ones conducted within CLEF evaluation forum. This year the task was performed for the second year on the biggest plant images dataset ever published in the literature. This dataset was composed of two distinct sources: a trusted set built from the GBIF and a noisy web dataset totaling both 4M images and covering 80k species. 

The main conclusion of our evaluation is that vision transformers performed definitely better than convolutional neural networks, especially when this type of models are pre-trained with a Self-Supervised Learning. Furthermore, an important lesson we have learned is the significance of maximizing the number of images, including those obtained from the web, despite the possibility of errors. It is crucial not to limit the size of the dataset based on organ types or assume that a certain number of training images is too small to be included in the test set. By incorporating a larger and more diverse set of images, we enhance the model's ability to capture a wider range of plant variations and improve its overall performance.

However, training those models requires more computational resources that only participants with access to large computational clusters can afford. For instance, the winning team Mingle Xu indicate that, they need to use 16 RTX 3090 GPUs for almost three months for training all the models. We are aware that this is not fair to other teams who do not have enough GPUs, and that it considerably limits the participation of other teams. However, we hope that the challenge and results presented in this article will highlight future research directions for solving key species identification problems across all kingdoms and advancing AI in general for biodiversity.

\section{Acknowledgments}
The research described in this paper was partly funded by the European Commission via the GUARDEN and MAMBO projects, which have received funding from the European Union’s Horizon Europe research and innovation program under grant agreements 101060693 and 101060639. The opinions expressed in this work are those of the authors and are not necessarily those of the GUARDEN or MAMBO partners or the European Commission.

\end{document}